\documentclass[letterpaper]{article} 
\usepackage{aaai2026}  
\usepackage{times}  
\usepackage{helvet}  
\usepackage{courier}  
\usepackage[hyphens]{url}  
\usepackage{graphicx} 
\usepackage{natbib}  
\usepackage{caption} 
\usepackage{algorithm}
\usepackage{algorithmic}

\usepackage{newfloat}
\usepackage{listings}
\DeclareCaptionStyle{ruled}{labelfont=normalfont,labelsep=colon,strut=off} 
\floatstyle{ruled}
\newfloat{listing}{tb}{lst}{}
\floatname{listing}{Listing}
\newcommand{\method}{\texttt{TAdaRAG}}
\newcommand{\benchmark}{\texttt{NowNewsQA}}
\usepackage{amsmath,amssymb,amsfonts}
\usepackage{subfig}
\usepackage{arydshln}
\usepackage[shortlabels]{enumitem} 
\usepackage{xcolor}
\usepackage{booktabs}
\usepackage[most]{tcolorbox}
\usepackage{tabularx}    
\usepackage{colortbl}  
\usepackage{multirow}
\title{TAdaRAG: Task Adaptive Retrieval-Augmented Generation via On-the-Fly Knowledge Graph Construction}
\author{
    Jie Zhang\textsuperscript{\rm 1}\equalcontrib,
    Bo Tang\textsuperscript{\rm 2,3}\equalcontrib,
    Wanzi Shao\textsuperscript{\rm 1},
    Wenqiang Wei\textsuperscript{\rm 3},
    Jihao Zhao\textsuperscript{\rm 3,4},
    Jianqing Zhu\textsuperscript{\rm 5},\\
    Zhiyu Li\textsuperscript{\rm 3},
    Wen Xi\textsuperscript{\rm 5},
    Zehao Lin\textsuperscript{\rm 3},
    Feiyu Xiong\textsuperscript{\rm 3},
    Yanchao Tan\textsuperscript{\rm 1}\thanks{Corresponding author.}
}

\affiliations {
    \textsuperscript{\rm 1}College of Computer and Data Science, Fuzhou University, Fuzhou 350116, China\\
    \textsuperscript{\rm 2}AIDS and SIAR, University of Science and Technology of China, Suzhou 215123, China\\
    \textsuperscript{\rm 3}MemTensor (Shanghai) Technology Co., Ltd., Shanghai 201306, China\\
    \textsuperscript{\rm 4}School of Information, Renmin University of China, Beijing 100086, China\\
    \textsuperscript{\rm 5}China Haisum Engineering Co., Ltd., Shanghai 200031, China\\
    ee6r1c7@gmail.com, tangbo@mail.ustc.edu.cn, yctan@fzu.edu.cn
}

\usepackage{bibentry}

\begin{document}
\maketitle
\setlength{\floatsep}{4pt plus 4pt minus 1pt}
\setlength{\textfloatsep}{4pt plus 2pt minus 2pt}
\setlength{\intextsep}{4pt plus 2pt minus 2pt}
\setlength{\dbltextfloatsep}{3pt plus 2pt minus 1pt}
\setlength{\dblfloatsep}{3pt plus 2pt minus 1pt}
\setlength{\abovecaptionskip}{3pt}
\setlength{\belowcaptionskip}{2pt}
\setlength{\abovedisplayskip}{2pt plus 1pt minus 1pt}
\setlength{\belowdisplayskip}{2pt plus 1pt minus 1pt}

\begin{abstract}
Retrieval-Augmented Generation (RAG) improves large language models by retrieving external knowledge, often truncated into smaller chunks due to the input context window, which leads to information loss, resulting in response hallucinations and broken reasoning chains.
Moreover, traditional RAG retrieves unstructured knowledge, introducing irrelevant details that hinder accurate reasoning.
To address these issues, we propose \method, a novel RAG framework for on-the-fly task-adaptive knowledge graph construction from external sources. 
Specifically, we design an intent-driven routing mechanism to a domain-specific extraction template, followed by supervised fine-tuning and a reinforcement learning-based implicit extraction mechanism, ensuring concise, coherent, and non-redundant knowledge integration. 
Evaluations on six public benchmarks and a real-world business benchmark (\benchmark) across three backbone models demonstrate that \method~outperforms existing methods across diverse domains and long-text tasks, highlighting its strong generalization and practical effectiveness.
\end{abstract}

\begin{links}
    \link{Code}{https://github.com/IAAR-Shanghai/TAdaRAG}
\end{links}

\setlength{\floatsep}{4pt plus 4pt minus 1pt}
\setlength{\textfloatsep}{4pt plus 2pt minus 2pt}
\setlength{\intextsep}{4pt plus 2pt minus 2pt}
\setlength{\dbltextfloatsep}{3pt plus 2pt minus 1pt}
\setlength{\dblfloatsep}{3pt plus 2pt minus 1pt}
\setlength{\abovecaptionskip}{3pt}
\setlength{\belowcaptionskip}{2pt}
\setlength{\abovedisplayskip}{2pt plus 1pt minus 1pt}
\setlength{\belowdisplayskip}{2pt plus 1pt minus 1pt}

\section{Introduction}\label{sec:section1}

In recent years, large language models (LLMs) ~\cite{openai2024gpt4o,guo2025deepseek,qwq2024} have achieved significant breakthroughs in natural language processing, particularly in tasks such as text generation and question-answering systems~\cite{yang2024crag, laban-etal-2024-summary}. Despite these advances, LLMs often generate plausible but factually incorrect responses, known as hallucinations~\cite{maynez2020faithfulness, zhou2020detecting,feng2024don,sun2024redeep}. To mitigate hallucinations, retrieval-augmented generation (RAG) integrates external knowledge sources into LLMs, providing enriched and contextually grounded inputs for more accurate responses~\cite{jiang2023active}.

\begin{figure}[t]
\includegraphics[width=\columnwidth]{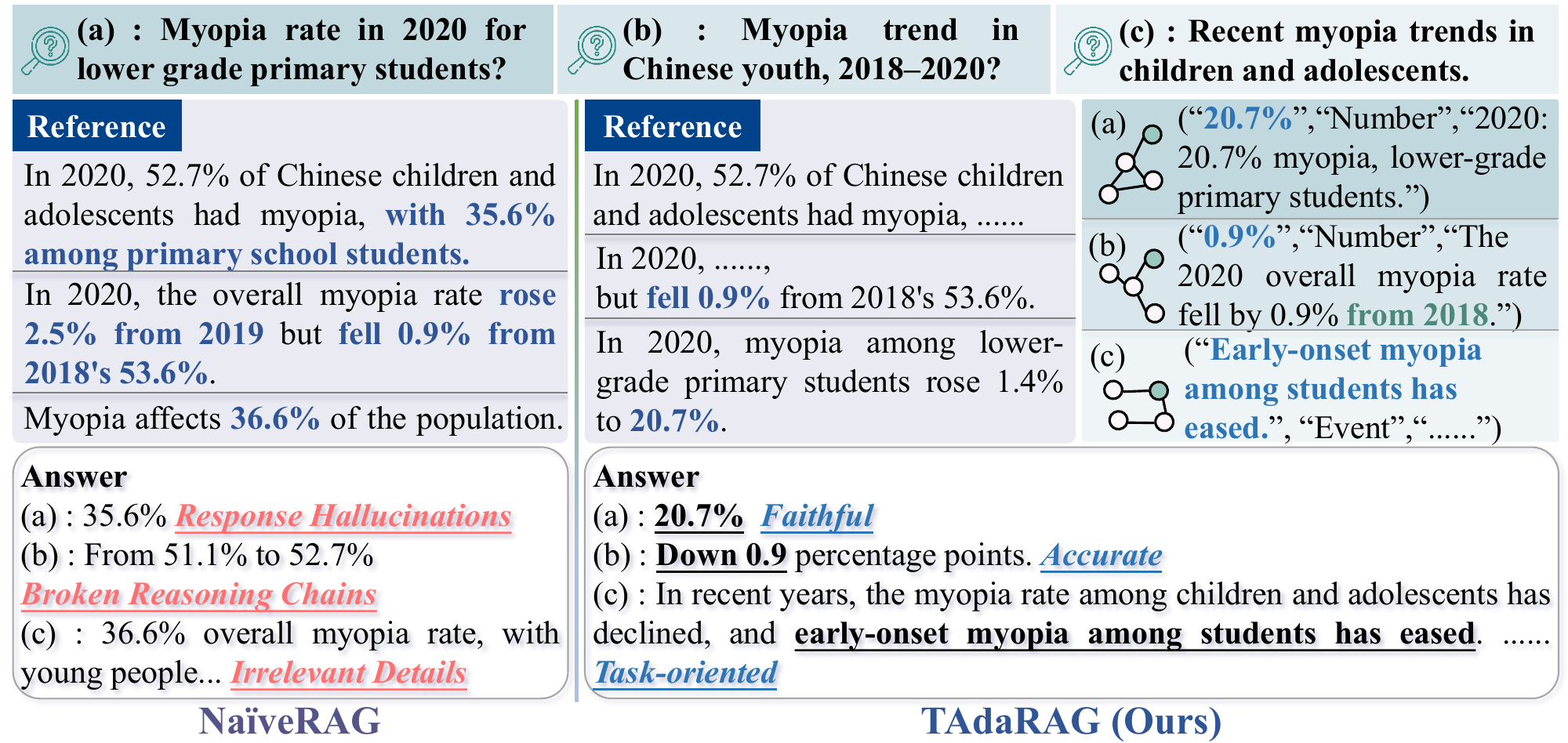}
  \caption{An illustrative example where the RAG system fails to generate correct responses due to truncation,
  leading to hallucinations, broken reasoning and irrelevant details. Our proposed \method~addresses these issues by integrating the task-adaptive knowledge graph dynamically.}
  \label{fig:toy_example}
\end{figure}

However, current RAG approaches face several critical limitations in practical scenarios, retrieving large amounts of relevant information, such as related documents, which are split into smaller chunks due to input size constraints~\cite{finardi2024chronicles,yepes2024financial}. 
This approach truncates complete knowledge, leading to information loss in chunks, which can cause hallucinations in responses, as shown in Case (a) of Figure \ref{fig:toy_example}.
Additionally, discrete chunks fail to capture the inherent logical relationships within the corpora, disrupting reasoning chains and affecting accuracy in complex tasks, as seen in Case (b).
Moreover, traditional RAG models input unorganized knowledge, recalling irrelevant details that hinder key information extraction and may impact practical usability, as illustrated in Case (c).

Recent graph-based RAG methods leverage knowledge graphs (KGs) to organize information through structured relationships, enhancing reasoning with summary-like responses ~\cite{Peng2024GraphRG, guo2024lightrag}.
However, these methods rely on preconstructed KGs, which require manual maintenance, lack scalability, and introduce redundant or incomplete information, limiting retrieval accuracy. 
To this end, we propose \method, a novel framework that diverges from traditional graph-based RAG by integrating task-oriented KG construction directly into the reasoning process, rather than during the retrieval phase. This dynamic structure mitigates text fragmentation hallucinations, enhances complex task reasoning, and enables more precise knowledge extraction without requiring predefined graphs.

Specifically, we first employ intent detection to route input texts to tailored extraction templates, ensuring precise initial graph construction across diverse domains. Then we refine the process through supervised fine-tuning, transforming fragmented external knowledge into concise, logically organized, and non-redundant structures. Finally, to achieve automatic KG extraction and self-optimization, we introduce an instruction-level implicit extraction mechanism optimized via reinforcement learning, which significantly enhances the relevance and accuracy of the extracted KGs.

We comprehensively evaluated the performance of \method~and existing RAG baselines on six public benchmarks and \benchmark, a business-scenario benchmark for Chinese current affairs news question answering (QA). Meanwhile, the incremental results at each stage shown in the ablation results demonstrate the effectiveness of our proposed method.
In addition, we conducted a human evaluation of the answers generated by different methods, and the results further confirm that \method~consistently outperforms other advanced baseline methods in terms of answer quality. We also examined the consistency between LLMs ratings and human ratings, as well as the consistency among expert ratings, further demonstrating the reliability of LLM scoring and \method.
To sum up, our main contributions are as follows:
\begin{itemize}
\item We propose \method, a task-adaptive RAG framework that integrates structured KG representations into the reasoning process. By dynamically constructing domain-relevant subgraphs, \method~effectively addresses the hallucination problem caused by chunked input in long-text tasks, enhances reasoning capabilities in complex scenarios, and ensures more accurate extraction and utilization of external knowledge.
\item We evaluate \method~on both public benchmarks and real-world business scenarios, covering tasks across various domains and long-text settings. The results show that \method~outperforms existing methods in multiple fields and long-context RAG tasks, demonstrating strong cross-domain generalization.
\item \method~has been successfully deployed in commercial applications, with trial accounts now available for user access and testing.
\end{itemize}

\section{Related Work}

\textbf{Retrieval-Augmented Generation (RAG).} RAG~\cite{borgeaud2022improving, 10.5555/3524938.3525306} enhances language models by retrieving information from external knowledge bases to improve text accuracy and credibility. Early methods encoded documents into vectors for fast retrieval~\cite{chen-etal-2024-m3, karpukhin-etal-2020-dense}, while later approaches introduced multi-step retrieval mechanisms to iteratively refine results and enhance long-text comprehension~\cite{jiang2023active, su-etal-2024-dragin, DBLP:conf/acl/TrivediBKS23}. RAG has been widely used in tasks such as Multi-Document QA~\cite{karpukhin2020dense, DBLP:conf/acl/TrivediBKS23}, summarization ~\cite{edge2024local, laban-etal-2024-summary}, and Open-domain QA~\cite{siriwardhana2023improving, yang2024crag, zhang-etal-2024-retrievalqa}. 
Recent advancements also explore adaptive retrieval strategies to better align retrieved content with query intent~\cite{mo2024aligning}. 
However, existing methods often process input text coarsely, introducing irrelevant information and omitting key details, which limits reasoning accuracy..

\textbf{Graph-enhanced Retrieval-Augmented Generation.} Graph-enhanced RAG~\cite{Peng2024GraphRG} models complex knowledge relationships through graph structures, enhancing retrieval comprehensiveness and reasoning capabilities. Early efforts extracted facts from predefined KGs (e.g., Wikidata~\cite{vrandevcic2014wikidata}) but struggled with dynamic task adaptation. Recent studies shift toward building task-specific graphs from raw text.
For example, GraphRAG~\cite{edge2024local} constructs KGs from textual data and generates community-based summaries; PathRAG~\cite{chen2025pathrag} significantly improves retrieval efficiency and reduces redundancy by identifying key relation paths; HippoRAG~\cite{jimenez2024hipporag} uses LLMs to convert document corpora into open KGs, serving as its artificial hippocampal index; and Chain of Knowledge~\cite{wang-etal-2024-boosting-language} helps mitigate hallucinations through structured evidence generation and rigorous verification.
These advancements highlight the potential of graph-based methods to dynamically adapt to varied tasks and improve contextual relevance. These studies underscore the role of graph structures in improving generation logic and reasoning accuracy. However, most existing methods rely on manually crafted graph models or static corpora, limiting their generalization ability.

\textbf{Handling Long-Context Tasks in LLMs.} In long-text tasks, LLM enhancement methods can be broadly divided into three categories. The first category focuses on expanding the context window through direct extension techniques~\cite{jin2024llm} and key-value (KV) cache pruning~\cite{zhang2023h2o}, where THINK~\cite{xu2024think} reduces memory overhead by pruning redundant KV channels based on low-rank attention patterns. 
Neurocache~\cite{safaya2024neurocache} introduces an external vector cache that stores compressed past states and employs efficient k-Nearest Neighbors retrieval to extend effective context lengths without full model retraining.
The second category improves long-text response quality through model fine-tuning, employing strategies such as supervised fine-tuning (SFT)~\cite{chen2023longlora} and reinforcement learning from human feedback (RLHF)~\cite{zhao2024context}. 
The third category introduces auxiliary structures to guide reasoning. For example, Quiet-STaR~\cite{zelikman2024quiet} enhances CoT reasoning by generating token-level predictions; MEMORAG~\cite{qian2024memorag} utilizes a memory module to generate retrieval cues; and INFO-RAG~\cite{xu2024unsupervised} refines retrieved content through a document optimizer. These methods help reduce noise, redundancy, and coherence issues in text generation but often require manually designed schemas or costly retraining.

\section{Methodology}
\begin{figure}[t]
  \centering
\includegraphics[width=\columnwidth]{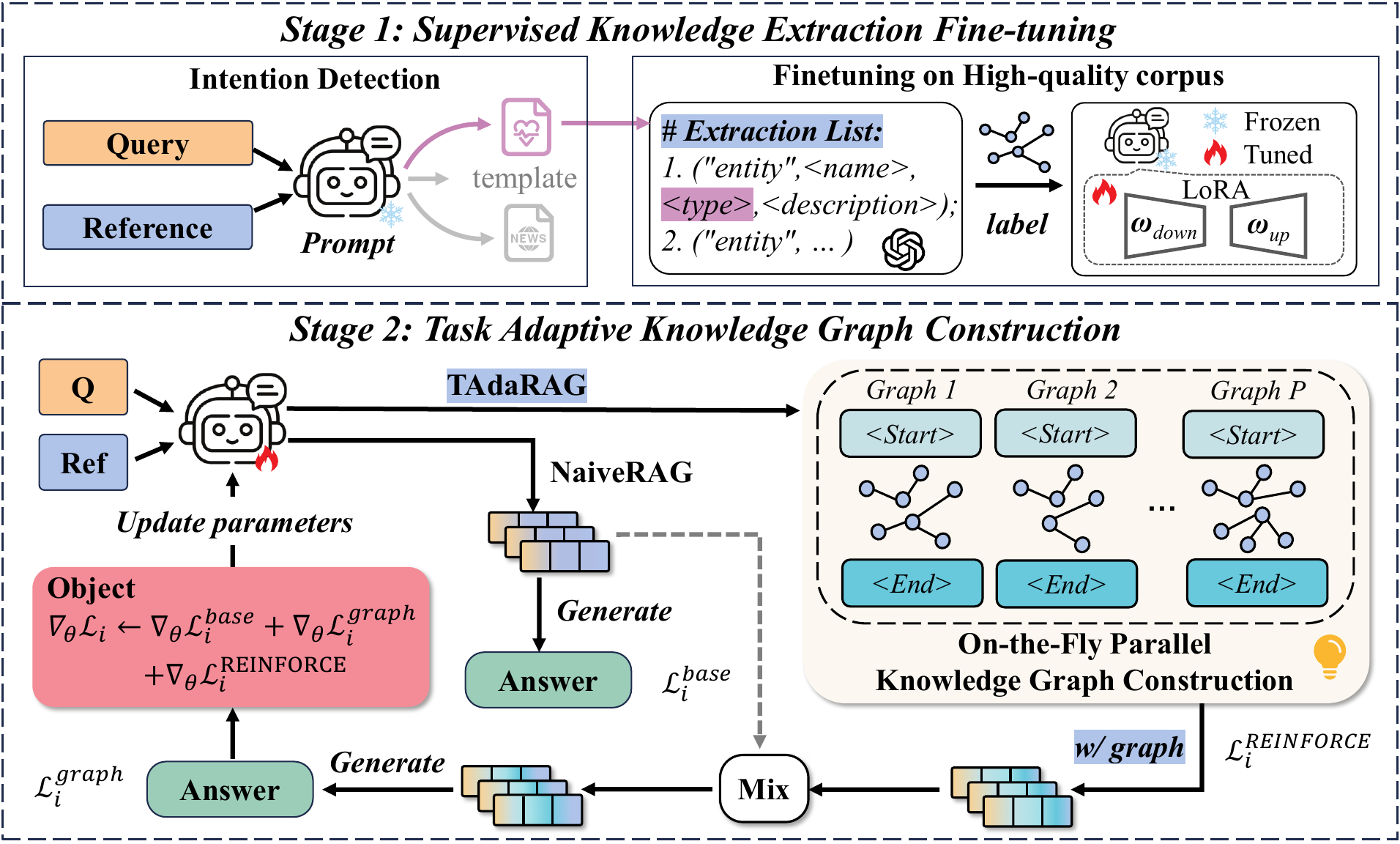} 
  \caption {An illustration of our proposed \method~framework and its two-stage training: (1) Supervised Knowledge Extraction Fine-tuning and (2) Task-Adaptive Knowledge Graph Construction.}
  \label{fig:framework}
\end{figure}

The \method~framework for language generation task (e.g., summarization) is formulated as follows. Given a database 
\(\mathbb{D} = \{(x_i, y_i)\}_{i=1}^{|\mathbb{D}|}\), where each pair \((x, y)\) represents a document and its summary, the framework consists of training and inference stages. 
For training (Figure \ref{fig:framework}), the \textbf{Supervised Knowledge Extraction Fine-tuning Stage} leverages strong LLMs ~\cite{openai2024gpt4o,guo2025deepseek} and domain-specific templates to generate KGs from \(\mathbb{D}\), enabling SFT for cold-start. The \textbf{Task-Adaptive Knowledge Graph Construction Stage} then trains the model to dynamically extract task-adaptive KG using the REINFORCE algorithm. 
During inference, the adaptive KG supports LLMs in generating accurate responses.

\subsection{Supervised
Knowledge Extraction Fine-tuning }\label{sec:subsection3.1}
\subsubsection{\textbf{Intention Detection}}
Pretrained language models often struggle with precise entity extraction, introducing irrelevant or redundant entities, which harms downstream performance, especially in real-world industry scenarios where capturing relevant entities is crucial for reliable responses.

To address this challenge, we introduce carefully designed external templates to guide entity extraction, reducing redundancy and improving entity relevance. Considering the diverse industry-specific requirements, we first identify key application domains and manually select high-impact entity types that most effectively support answer generation. Additionally, we design tailored extraction templates for general as well as specialized scenarios, enabling the model to determine the required entity types, entity description specifications, and relationships among entities for each domain. This significantly enhances the model's cross-domain generalization capability. 

In practice, given a user query \( q \) and external knowledge \( r \), we use prompts to perform intent detection and select the appropriate template \( t \), enabling the model to accurately identify the required node types and relation patterns for the current task. The model then extracts a typed knowledge graph that directly aligns with the detected intent and maximizes downstream answer quality. In the RAG setting, this dynamically constructed, domain-aware knowledge graph is integrated into the generation pipeline to produce more accurate and contextually appropriate responses.

\subsubsection{\textbf{Fine-tuning on High-Quality Corpus}}
Based on the obtained template \( t \), we construct an instruction set \( I = \{q, r, t\} \) by integrating the question \( q \) and external knowledge \( r \), serving as data samples for supervised knowledge extraction fine-tuning~\cite{friel2024ragbench}.  
Then, we use strong LLMs to perform the knowledge extraction task on the instruction set, generating high-quality knowledge extraction results \( G \).
Based on these instruction-graph pairs, we create a high-quality dataset for supervised knowledge extraction fine-tuning, covering four question domains and seven sub-datasets, totaling 9,548 fine-tuning samples.
Finally, leveraging LoRA ~\cite{hu2022lora} for supervised fine-tuning on a pretrained LLM with the dataset, we train the model  to achieve excellent knowledge extraction capabilities, laying a crucial foundation for on-the-fly construction of high-quality KGs.

\subsection{Task-Adaptive Knowledge Graph Construction }\label{sec:subsection3.2}
\subsubsection{\textbf{Parallel Construction}}
LLMs' inability to assess single KG quality can lead to suboptimal outcomes when relying on only one KG for next-token generation.
Sampling multiple candidate KGs helps mitigate this issue.
Therefore, based on the training set \( \mathbb{D}_{train} = \{(x_i, y_i)\}_{i = 1}^{N} \), the model performs knowledge extraction on the input text and constructs \( p \) parallel subgraphs \( g_i = \{g_i^1, g_i^2, \dots, g_i^p \} \).
Learnable tokens \texttt{<|startextraction|>} and \texttt{<|endextraction|>} are used to indicate the start and end positions of knowledge extraction, enabling the model to naturally embed the knowledge graph during generation and achieve implicit knowledge extraction.
\subsubsection{\textbf{Mixing Network}}
To enhance the model's understanding and utilization of structured information in knowledge graphs, we propose a graph-structured fusion network. Given an instruction-response pair \( (x_i, y_i) \) and a subgraph \( g_i^k \), we first obtain the hidden state \( H_{i,j}^{\text{base}} = \text{hs}[x_i; y_i^{(j-1)}] \) of the \( j \)-th token in the response under the original RAG method. In our proposed \method, which incorporates the subgraph \( g_i^k \), the corresponding hidden state is \( H_{i,j,k}^{\text{graph}} = \text{hs}[x_i; y_i^{(j-1)}; g_i^k] \).

To better measure the importance of knowledge graph compared to direct responses, we concatenate the two obtained hidden states and input them into a three-layer MLP with ReLU activation to compute the weight for each token in the response:
\begin{align}
\omega_{i,j,k} = \text{MLP}(\text{concat}(H_{i,j}^{\text{base}}, H_{i,j,k}^{\text{graph}})).
\end{align}
where concat(·) denotes the concatenation.

Meanwhile, we compute the log-likelihoods under the conditions of without and with the knowledge graph:
\begin{align}
l_{i,j,k}^{\text{w/o graph}} &= \log p_{\theta} (y_i^j \mid x_i, y_i^{(j-1)})  = f(H_{i,j}^{\text{base}}),
\end{align}
\begin{align}
l_{i,j,k}^{\text{w/ graph}} &= \log p_{\theta} (y_i^j \mid x_i, y_i^{(j-1)}, g_i^k) = f(H_{i,j,k}^{\text{graph}}),
\end{align}
where \( f(\cdot) = \text{softmax}(\text{lmhead}(\cdot)) \), \( \text{lmhead}(\cdot) \) maps the hidden state to the vocabulary space to obtain logits, and \( \text{softmax}(\cdot) \) converts them into a probability distribution.

Finally, we compute the weighted sum of the \( j \)-th token to obtain the log-likelihood incorporating the graph structure for generation and optimization:
\begin{equation}
l_{i,j,k}^{\text{mix}} = \omega_{i,j,k} \cdot l_{i,j,k}^{\text{w/ graph}} + (1 - \omega_{i,j,k}) \cdot l_{i,j,k}^{\text{w/o graph}}.
\end{equation}
\subsubsection{\textbf{Optimizing Graph Construction}}
Given that the extracted KGs vary in length and content, we aim to determine the most beneficial subgraph for optimizing the response. Specifically, for all instruction-answer pairs \( (x_i, y_i) \) in the training set \( \mathbb{D}_{\text{train}} \), our objective is to find an optimal subgraph \( \tilde{g}^{(i)} \) that maximizes \(
\pi_{\theta} (y_i \mid x_i, \tilde{g}^{(i)}) 
\).

To achieve this, we design a reward function \( R \) based on the REINFORCE algorithm to quantify the impact of introducing subgraphs on response generation.
We first consider the loss of model direct response to ensure its ability to answer without external knowledge:
\begin{equation}
\mathcal{L^{\text{base}}} = -\mathbb{E}_{(x, y) \sim \mathbb{D}} \left[ \log \pi_{\theta} (y \mid x) \right].
\end{equation}

Next, we consider the loss when incorporating the knowledge graph into the response, aiming to train the model to learn how to integrate the input instruction and subgraph to generate more accurate answers:
\begin{equation}
\mathcal{L^{\text{graph}}} = - \mathbb{E}_{(x, y, g) \sim \mathbb{D} \cup g} \left[ \log \pi_{\theta} (y \mid x, g) \right].
\end{equation}

At this point, for the case where the \(i\)-th instruction references the \(k\)-th subgraph, we design the reward function \( R \) as follows:
\begin{equation}
R_{i,k} = \max(0, \mathcal{L}_i^{\text{base}} - \mathcal{L}_{i,k}^{\text{graph}} - \bar{R}_i).
\end{equation}
This reward function increases the likelihood of selecting knowledge graphs that perform better than the average $\bar{R}_i$. Finally, the REINFORCE loss term is defined as:
\begin{equation}
\mathcal{L^{\text{REINFORCE}}} = - R_{i,k} \cdot \log \pi_{\theta} (g_i^k \mid x_i).
\end{equation}

Thus, the total loss function for model training is defined as:
\begin{equation}
\mathcal{L} = \alpha \cdot \mathcal{L^{\text{base}}} + (1 - \alpha) \cdot \mathcal{L^{\text{graph}}} + \beta \cdot \mathcal{L^{\text{REINFORCE}}},
\end{equation}
where \( \alpha \) and \( \beta \) are hyperparameters.

\section{Experiments}\label{sec:section4}
\begin{table*}[t]
\centering
\small
\begin{tabularx}{\textwidth}{c c|*{3}{>{\centering\arraybackslash}X}|*{3}{>{\centering\arraybackslash}X}}
\toprule
\multicolumn{2}{c|}{\multirow{2}{*}{\centering Methods}} & \multicolumn{3}{c|}{ULTRADOMAIN} & \multicolumn{3}{c}{LongBench} \\
\cmidrule(lr){3-5} \cmidrule(lr){6-8}
 & & Health & Biology & Legal & HotpotQA & 2WikiMQA & GovReport \\
\midrule
\multicolumn{8}{c}{\cellcolor[HTML]{EFEFEF}\textbf{Based on Mistral-7B-Instruct}} \\
\midrule
\multirow{2}{*}{Standard RAG} & \multicolumn{1}{|c|}{NaïveRAG} & 34.80 & 34.10 & 35.80 & 37.60 & 20.60 & 27.40 \\
& \multicolumn{1}{|c|}{BGE-M3}   & 33.20 & 32.20 & 42.00 & 36.20 & 20.30 & 26.10 \\

\midrule
\multirow{4}{*}{Advanced RAG}
  & \multicolumn{1}{|c|}{RQ-RAG}      & 33.37 & 33.42 & 42.60 & 37.00 & 21.50 & 18.60 \\
  & \multicolumn{1}{|c|}{GraphRAG}        & 35.60 & 34.80 & 37.65 & 38.00 & 36.50 & 25.60 \\
  & \multicolumn{1}{|c|}{HippoRAG}        & 34.54 & 34.23 & 35.36 & 39.30 & 33.10 & 25.22 \\
  & \multicolumn{1}{|c|}{PathRAG}     & 21.67 & 20.10 & 18.57 & 24.22 & 18.71 & 15.66 \\
  & \multicolumn{1}{|c|}{MEMORAG}     & 37.40 & 35.70 & \textbf{51.20} & \underline{42.90} & 30.30 & 31.60 \\

\midrule
\multirow{3}{*}{TAdaRAG}
  & \multicolumn{1}{|c|}{w/ graph}      & 38.19 & 36.87 & 32.92 & 38.30 & 38.48 & 33.72 \\
  & \multicolumn{1}{|c|}{w/ sft}        & \underline{40.00} & \underline{38.92} & 39.32 & 41.60 & \underline{38.86} & \underline{35.39} \\
  & \multicolumn{1}{|c|}{w/ reinforce}  & \textbf{40.77}\textsuperscript{*} & \textbf{39.31}\textsuperscript{*} & \underline{49.88} & \textbf{44.83}\textsuperscript{*} & \textbf{39.31}\textsuperscript{*} & \textbf{36.41}\textsuperscript{*} \\
  
\midrule
\multicolumn{8}{c}{\cellcolor[HTML]{EFEFEF}\textbf{Based on Qwen2.5-7B-Instruct}} \\
\midrule

\multirow{2}{*}{Standard RAG}
  & \multicolumn{1}{|c|}{NaïveRAG}    & 35.25 & 35.28 & 36.55 & 45.78 & 32.28 & 20.68 \\
  & \multicolumn{1}{|c|}{BGE-M3}      & 30.20 & 33.20 & 40.60 & 36.28 & 33.30 & 20.1 \\

\midrule
\multirow{4}{*}{Advanced RAG}
  & \multicolumn{1}{|c|}{RQ-RAG}      & 31.50 & 31.90 & 38.80 & 37.40 & 34.10 & 21 \\
  & \multicolumn{1}{|c|}{GraphRAG}        & 36.82 & 34.67 & 40.62 & 43.33 & 37.52 & 28.46 \\
  & \multicolumn{1}{|c|}{HippoRAG}        & 35.73 & 35.53 & 40.31 & 45.89 & 36.16 & 27.23 \\
  & \multicolumn{1}{|c|}{PathRAG}    &32.65  &30.46  &32.34  &33.79  &29.09  &24.55  \\
  & \multicolumn{1}{|c|}{MEMORAG}     & 36.87 & 36.00 & \textbf{47.60} & 37.99 & 35.32 & 31.13 \\

\midrule
\multirow{3}{*}{TAdaRAG}
  & \multicolumn{1}{|c|}{w/ graph}      & 40.77 & 38.31 & 41.76 & 48.74 & 42.84 & 33.88 \\
  & \multicolumn{1}{|c|}{w/ sft}        & \underline{41.35} & \underline{39.62} & 44.55 & \underline{49.03} & \underline{43.37} & \underline{35.47} \\
  & \multicolumn{1}{|c|}{w/ reinforce}  & \textbf{42.38}\textsuperscript{*} & \textbf{40.75}\textsuperscript{*} & \underline{46.83} & \textbf{49.23}\textsuperscript{*} & \textbf{43.79}\textsuperscript{*} & \textbf{36.95}\textsuperscript{*} \\
  
\bottomrule
\end{tabularx}
\caption{Experiments based on Mistral-7B-Instruct and Qwen2.5-7B-Instruct. GovReport uses the ROUGE-L metric, and other datasets use the F1 metric. The w/ graph variant leverages KGs via prompting, w/ sft enhances knowledge extraction through SFT, and w/ reinforce optimizes task-adaptive KG construction with reinforcement learning. * indicates statistically significant improvements ($p < 0.01$) over SOTA RAG baselines.}
\label{tab:results}
\end{table*}
\begin{figure*}[t]
  \centering
  \begin{minipage}[t]{0.48\textwidth}
    \centering
    \includegraphics[width=\linewidth]{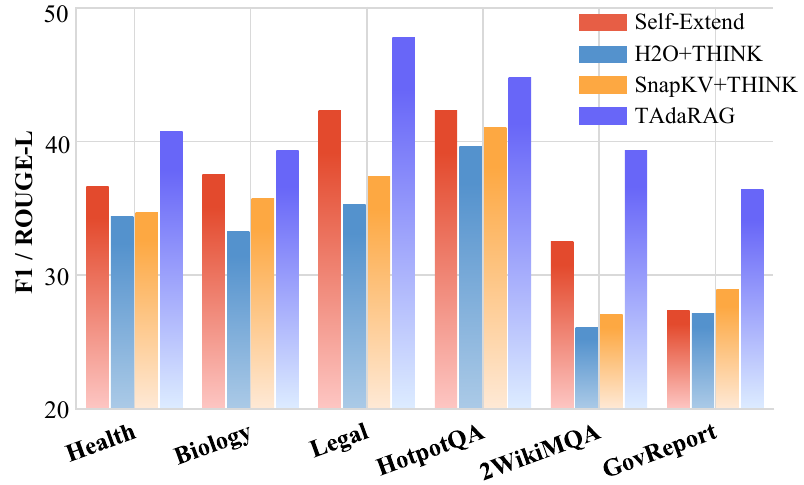}
    \caption{Long-context experiments on Mistral-7B.}
    \label{fig:long-text}
  \end{minipage}%
  \hfill
  \begin{minipage}[t]{0.48\textwidth}
    \centering
    \includegraphics[width=\linewidth]{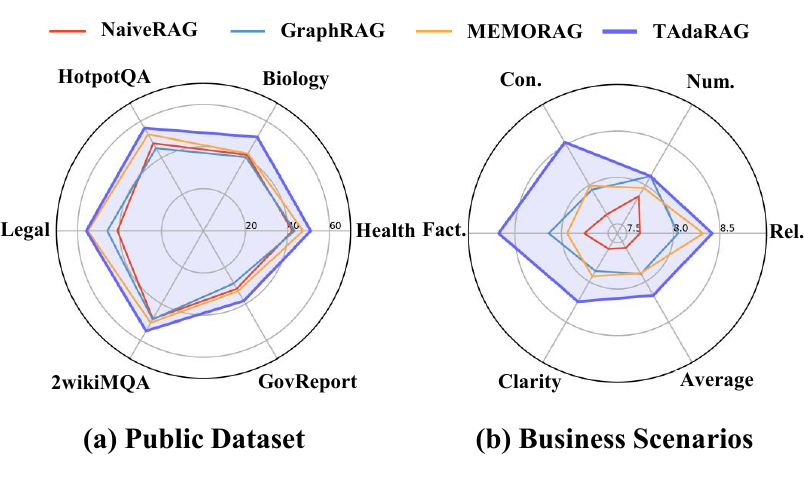}
    \caption{Experiments on Qwen2.5-14B.}
    \label{fig:qwen14b}
  \end{minipage}
\end{figure*}
In the experiments, we investigated the following research questions (\textbf{RQs}): 
\textbf{RQ1:} How does the effectiveness of our proposed \method~framework compare to the state-of-the-art RAG baselines?
\textbf{RQ2:} How effectively does the \method~framework handle long-context tasks? 
\textbf{RQ3:} How does the \method~framework generalize to real-world task scenarios? 
\textbf{RQ4:} Has each training stage of our framework contributed effectively?
\textbf{RQ5:} How do different hyperparameter values in \method~influence the performance?

In addition to addressing these questions, we conducted further experiments including statistical significance testing, latency analysis, KG refinement evaluation, evidence verification and case studies in the extend version.

\subsection{Evaluation of Public Datasets}\label{sec:subsection4.1}

\textbf{Datasets \& Metrics.} 
(1) \textbf{For Q\&A Tasks}: We conducted experiments on Health, Biology, and Legal datasets for open-domain QA~\cite{qian2024memorag}, and HotpotQA and 2WikiMQA for restricted QA~\cite{yang2018hotpotqa, ho2020constructing, bai2024longbench}, evaluated with F1. 
(2) \textbf{For Summarization}: We used GovReport dataset to evaluate summary generation~\cite{huang2021efficient}, evaluated with ROUGE-L. 

\textbf{Baseline.}
We compare \method~against seven representative RAG approaches. 
Na\"iveRAG~\cite{gao2023retrieval} implements the standard RAG paradigm. 
BGE-M3~\cite{chen2024bge} is a versatile text embedding model with multi-lingual, multi-task, and multi-granularity capabilities. 
RQ-RAG~\cite{chan2024rq} extends this framework by jointly learning to reformulate queries before retrieval, thereby improving the relevance of retrieved passages. 
GraphRAG~\cite{edge2024local} adopts LLM-extracted knowledge graphs and community detection to enable scalable, comprehensive summarization for complex queries.
HippoRAG~\cite{jimenez2024hipporag} adopts hippocampal indexing theory and KGs for reasoning.
MEMORAG~\cite{qian2024memorag} effectively expands the application scope of the RAG system by introducing memory modules. Finally, 
PathRAG~\cite{chen2025pathrag} retrieves nodes, prunes key paths, and converts them to text to guide LLM generation. 

\textbf{Implementation details.}
In this study, we adopt Mistral-7B-Instruct~\cite{mistral7b} , Qwen2.5-7B-Instruct and Qwen2.5-14B-Instruct~\cite{qwen2.5} as the backbone model. For the two-stage training strategy, we perform SFT in \textbf{Stage 1} for 5 epochs, with a maximum input sequence length of 20{,}480 tokens, a batch size of 1, gradient accumulation over 8 steps, and a cosine learning rate scheduler initialized at 5e\textminus5. 
In \textbf{Stage 2}, we train the model using ZeRO stage-2 optimization with the AdamW optimizer, a per-GPU batch size of 1, and bfloat16 precision, for 3 epochs at a learning rate of 5e\textminus7. During training, we generate multiple KGs per instruction using a sampling temperature $T=0.6$; for evaluation, we apply greedy decoding. The maximum KG length is set to 2048 tokens, with longer outputs truncated. The entire training process takes approximately 16 hours using 8 NVIDIA A100 (80~GB) GPUs, with 4 hours for Stage~1 and 12 hours for Stage~2.

\subsubsection{\textbf{Main Results (RQ1)}}
Table \ref{tab:results} and Figure \ref{fig:qwen14b} highlights several key insights. Taking Mistral-7B results as an example:

\textbf{Hallucination Mitigation.} 
\method~mitigates information loss in chunked text through on-the-fly knowledge graph construction with two-stage training, outperforming state-of-the-art (SOTA) RAG baselines MEMORAG in factual domains (Health: 37.40 → 40.77; Biology: 35.70 → 39.31). 
On the Legal dataset, \method~significantly outperforms NaiveRAG (35.80 → 49.88), demonstrating strong factuality in answering questions that require the integration of lengthy legal clauses. Its close performance to MEMORAG (49.88 vs. 51.20) further suggests potential for domain-specific enhancement.
These results demonstrate that the structured knowledge integration of \method~not only improves answer accuracy but also enhances model robustness across domains.

\textbf{Reasoning Enhancement.} 
\method~enhances reasoning chain completeness by dynamically organizing knowledge hierarchies. Compared to MEMORAG, it achieves notable improvements on complex reasoning tasks (2WikiMQA: 30.30 → 39.31) and multi-hop question answering (HotpotQA: 42.90 → 44.83), demonstrating its superior ability to support structured, multi-step inference.

\textbf{Task-Oriented Extraction.} 
\method~outperforms the SOTA MEMORAG in summarization tasks (GovReport: 31.60 → 36.41), indicating improved scalability and precision in integrating task-oriented knowledge for long-text summarization.

\subsubsection{\textbf{Long-Context Task Analysis (RQ2)}}
We compared \method~with three long-context mechanisms on the same base model: Self-Extend ~\cite{jin2024llm}, H2O+THINK, and SnapKV+THINK ~\cite{xu2024think} on six datasets, using the same foundation model, as shown in Figure \ref{fig:long-text}.
On HotpotQA and 2WikiMQA, \method~effectively handles multi-document QA, rivaling dedicated long-context models without modifying storage. In GovReport, it excels in long-document summarization, demonstrating the benefits of task-adaptive KG construction for long-context task.

\begin{figure*}
  \centering
\includegraphics[width=1\textwidth]
  {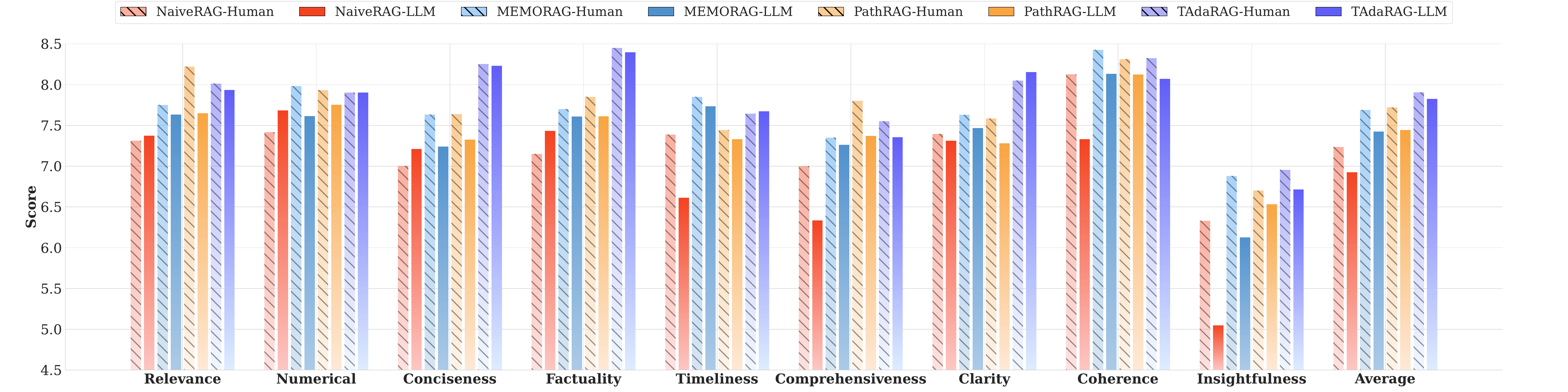}
  \caption{Multi-faceted comparison of different methods based on human and GPT-4o. Higher values indicate better performance, with a maximum value of 10.}
  \label{fig:human-vs-LLM}
\end{figure*}

\subsection{Evaluation of Business Scenarios}\label{sec:subsection4.2}

\textbf{Dataset Construction.} 
Based on real-world business scenarios from Xinyu AI Search ~\cite{wu2024xinyu}, we crafted \benchmark, a multi-document QA dataset focused on the news domain, consisting of 3,150 examples (3,000 for training and 150 for testing).
The questions were derived from diverse real-user queries covering trending topics, political developments, economic shifts, and major societal events, reflecting the dynamic and multifaceted nature of news consumption. 
Reference documents were retrieved using the Xinyu AI Search engine, which employs hybrid retrieval techniques to aggregate and rank documents based on their relevance to each query. 
Each document set mirrors actual search outputs from a production-level engine, preserving realistic characteristics such as redundancy, noise, and partial relevance. 
As such, \benchmark~serves as a challenging benchmark for evaluating model performance in reasoning over fragmented news content within real-world business contexts.

\textbf{Multi-faceted Evaluation Criteria.} 
Because it is difficult to establish a gold-standard answer for RAG tasks—especially given the inherent diversity and subjectivity of answers in real-world industry scenarios—we adopted a rating-based evaluation framework instead of relying on exact match metrics. 
To ensure the robustness and relevance of the evaluation, we invited domain experts with journalism backgrounds and master's degrees to design a set of multi-dimensional criteria that comprehensively assess the quality of generated answers. 
The criteria include: (1) Relevance, (2) Numerical Precision, (3) Conciseness, (4) Factuality, (5) Timeliness, (6) Comprehensiveness, (7) Clarity, (8) Coherence, and (9) Insightfulness. 

\textbf{LLM Evaluation.} 
Conducting comprehensive human evaluations using multi-faceted criteria for all experiments is cost-prohibitive.
To address this, we adopted an LLM to assess generated answers. This automated evaluation strategy has been widely applied in prior work~\cite{chen2025pathrag, Peng2024GraphRG, qian2024memorag}.
Specifically, we utilized GPT-4o, prompting it to generate point-by-point justifications and assign a final score based on the defined criteria. The temperature was set to 0 to ensure deterministic outputs. 
As shown in Table~\ref{tab:pearson}, the LLM-based scores exhibit strong correlation with human evaluations, as evidenced by the comparative results in Figure~\ref{fig:human-vs-LLM}.

\begin{table}[H]
  \centering
  \small
  \begin{tabular}{l@{\hskip 6pt}r@{\hskip 12pt}l@{\hskip 6pt}r}
    \toprule
    \textbf{Metric} & \textbf{Value} & \textbf{Metric} & \textbf{Value} \\
    \midrule
    Relevance         & 0.706 & Comprehensiveness & 0.850 \\
    Numerical Precision & 0.755 & Clarity           & 0.867 \\
    Conciseness       & 0.847 & Coherence         & 0.925 \\
    Factuality        & 0.842 & Insightfulness    & 0.881 \\
    Timeliness        & 0.828 & --                & --    \\
    \bottomrule
  \end{tabular}
    \caption{Pearson correlation coefficients between human and LLM scores for different evaluation criteria.}
    \label{tab:pearson}
\end{table}

\textbf{Baseline.}
We test \method~against Na\"iveRAG~\cite{gao2023retrieval}, MEMORAG~\cite{qian2024memorag}, and PathRAG~\cite{chen2025pathrag} on real-world business scenarios based on Mistral-7B-Instruct, with performance assessed under multi-faceted evaluation criteria.

\subsubsection{\textbf{Comparative Evaluation (RQ3)}}

\textbf{Multi-Faceted Evaluation of the Generated Answer.}
We compare \method~with existing RAG systems by inviting human experts to assess the generated answers using the multi-faceted evaluation criteria. As shown in Figure~\ref{fig:human-vs-LLM}, \method~achieves the highest overall average score (7.904 vs. 7.720), demonstrating superior performance across multiple dimensions. Notably, it significantly outperforms other methods in conciseness (8.251 vs. 7.637) and factuality (8.449 vs. 7.850). These results underscore the effectiveness of our task-adaptive knowledge graph construction in mitigating hallucinations and enhancing response accuracy. Furthermore, \method's strong performance on real-world business queries highlights its practical feasibility and broad generalizability in applied settings.

\textbf{LLM-Based Multi-Faceted Evaluation. }
Figure~\ref{fig:human-vs-LLM} reports the results of the multi-faceted evaluation conducted by GPT-4o. Although the absolute scores differ from those given by human annotators (see Figure~\ref{fig:human-vs-LLM}, the relative rankings are largely consistent, indicating a strong correlation between the two (see Table~\ref{tab:pearson}). These findings suggest that LLM-based evaluation is a reliable proxy for human judgment and can effectively reflect model performance across key dimensions.

\begin{table*}[t]
  \centering
  \small
  \begin{tabularx}{\textwidth}{l*{9}{>{\centering\arraybackslash}X}}
    \toprule
    Model & Rel. & Num. & Concise & Fact. & Time. & Comp. & Clarity & Coh. & Insight  \\
    \midrule
    Na\"iveRAG & 0.813 & 0.727 & 0.777 & 0.814 & 0.742 & 0.663 & 0.774 & 0.764 & 0.676 \\
    MEMORAG & 0.764 & 0.701 & 0.763 & 0.841 & 0.747 & 0.667 & 0.815 & 0.722 & 0.691 \\
    PathRAG & 0.797 & 0.674 & 0.807 & 0.806 & 0.759 & 0.716 & 0.712 & 0.691 & 0.654 \\
    TAdaRAG & 0.791 & 0.691 & 0.856 & 0.821 & 0.719 & 0.683 & 0.806 & 0.68 & 0.694 \\
    \bottomrule
  \end{tabularx}
  \caption{Pearson correlation coefficients for human scoring consistency verification.}
  \label{tab:expert}
\end{table*}
\textbf{Human Scoring Consistency Verification. }
To verify the consistency of human judgments, we computed the inter-rater agreement among three expert annotators on 150 test samples using Pearson correlation across nine evaluation dimensions. The average correlation scores for each metric are reported in Table~\ref{tab:expert}. As shown, all models exhibit consistently high agreement, indicating that the evaluation criteria are well-defined and easy to apply. These strong correlations further suggest that human ratings are stable and reliable, thereby reinforcing the credibility of the evaluation results presented in this study.

\subsection{Ablation Study (RQ4)}\label{sec:subsection4.3}
\begin{figure}
  \centering
\includegraphics[width=\columnwidth]
  {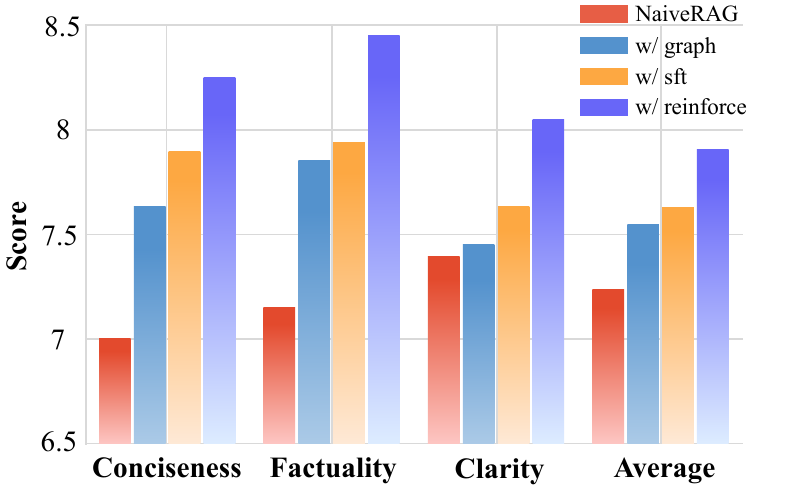}
  \caption{Ablation study on \benchmark.}
  \label{fig:bar_chart}
\end{figure}
This section investigates the impact of structured KGs in our setting and validates the necessity of the two-stage training design in \method. The results, summarized in Table~\ref{tab:results} and Figure~\ref{fig:bar_chart}, reveal that:

\textbf{The introduction of prompt-based KG}, which utilizes KGs extracted by the model itself through domain-specific templates, significantly improves model performance across various datasets.
For example, it outperforms Na\"iveRAG, with 2WikiMQA improving from 20.60 to 38.48 and GovReport from 27.40 to 33.72, indicating that KGs effectively enhance model reasoning and summarization capabilities.
Moreover, the results of direct prompt-based integration validate the feasibility of enhancing the model through KG incorporation, providing a solid basis for the subsequent optimization of KG construction.

\textbf{Supervised Knowledge Extraction Fine-tuning} enables the model to achieve ideal KGs extraction, bridges prompt-based KG construction and task adaptation.
For instance, it improves accuracy by 19.44\% on Legal dataset and by 5.56\% on Biology dataset, demonstrating its effectiveness in optimizing knowledge extraction and boosting answer precision. 
Results outperform the prompt-based version, indicating that higher-quality KGs lead to better performance.

\textbf{Task-Adaptive KG Construction} further improves the model across all datasets and tasks, validating the effectiveness of our REINFORCE algorithm. 
For example, it significantly enhances complex question answering in the Legal domain, with performance improving by 26.86\% compared to Stage 1. 
Moreover, it boosts results across other datasets to achieve SOTA performance, showcasing the indispensability of this stage.

\subsection{Hyperparameter Analysis (RQ5)}\label{sec:subsection4.4}
\begin{figure}
  \centering
\includegraphics[width=\columnwidth]
  {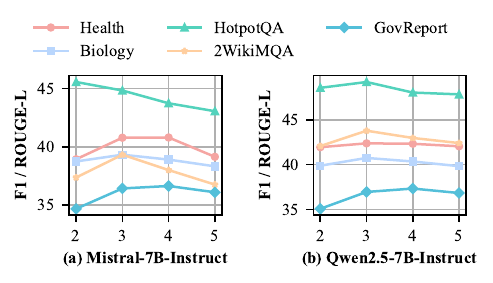}
  \caption{Parallel Subgraphs vs. Performance}
  \label{fig:line_chart}
\end{figure}
We investigate the impact of varying the number of parallel subgraphs sampled during training on model performance. 
As shown in Figure \ref{fig:line_chart}, both Mistral-7B and Qwen2.5-7B achieve optimal performance when the number of subgraphs is set to 3 across most datasets. 
Using a subgraph number of 2 restricts the model’s reasoning capacity and impedes the identification of optimal substructures. Increasing the number beyond 3 (e.g., 4 or 5) may introduce noise into the optimization process for models with 7B parameters, thereby degrading performance.
Specifically, HotpotQA benefits from 2 subgraphs, as this suffices for focused reasoning, whereas GovReport performs better with 4 subgraphs due to its need for more diverse and informative subgraphs to support evidence aggregation in summarization.
Moreover, experiments with Qwen show stable performance across varying subgraph numbers, suggesting that stronger models with higher performance exhibit greater training stability.

\section{Conclusion}\label{sec:section5}
In this paper, we present \method, a task-adaptive RAG framework that dynamically constructs structured knowledge graphs to mitigate hallucinations, strengthen reasoning, and enhance knowledge extraction.
Extensive experiments on six public benchmarks and a real-world business dataset demonstrate consistent improvements across factual QA, multi-hop reasoning, long-text summarization, and industry scenarios, highlighting strong generalization and practical utility.
The current design involves dynamic KG construction and multi-stage training, adding computational overhead and relying partly on manually crafted templates, which may constrain efficiency and adaptability in more complex scenarios. 
Future work will further improve the efficiency of KG construction, reduce computation cost, and enhance adaptability, aiming to strengthen the framework’s scalability and reliability in real-world settings.


\section{Acknowledgments}
This work was supported in part by the Fujian Provincial Artificial Intelligence Industry Development Technology Project under Grant (2025H0042) and Talent Foundation of Fuzhou University (No. XRC-23027).

\bibliography{arxiv}

\clearpage

\appendix
\section{Appendix A: Data Resources}
\label{sec:appendix A}

\subsection{A1: Dataset Descriptions}
\label{sec:appendix A1}
We conducted experiments on six public datasets and a real-world business benchmark (denoted as \benchmark), with dataset statistics summarized in Table~\ref{tab:datasets}. For our task, which leverages KGs to enhance answer generation, we additionally report the typical range of graph sizes and the number of entities used across different datasets.

\begin{table*}[ht]
\centering
\begin{tabular}{@{}l|ccccccc@{}}
\toprule
\textbf{Datasets} & \textbf{Health} & \textbf{Biology} & \textbf{Legal} & \textbf{HotpotQA} & \textbf{2WikiMQA} & \textbf{GovReport} & \textbf{\benchmark} \\
\midrule
\textbf{Number of Samples} & 180 & 220 & 438 & 200 & 200 & 200 & 150 \\
\textbf{Metric} & F1 & F1 & F1 & F1 & F1 & ROUGE-L & Multi-Faceted \\
\textbf{Avg. Length} & 135{,}901 & 125{,}284 & 51{,}413 & 9{,}149 & 4{,}885 & 8{,}169 & 10{,}296 \\
\textbf{Avg. Entities} & 44.23 & 42.32 & 15.64 & 10.09 & 8.23 & 18.82 & 24.95 \\
\textbf{Avg. Graph Size} & 2005.87 & 2043.98 & 1877.84 & 256.67 & 298.21 & 1240.87 & 1887.01 \\
\bottomrule
\end{tabular}
\caption{Statistics of six public datasets and one real-world business benchmark.}
\label{tab:datasets}
\end{table*}

\begin{table*}
  \centering
  \begin{tabularx}{\linewidth}{l *{6}{>{\centering\arraybackslash}X}}
    \toprule
    \textbf{Model} & \textbf{Health} & \textbf{Biology} & \textbf{Legal} & \textbf{HotpotQA} & \textbf{2WikiMQA} & \textbf{GovReport} \\
        \midrule
\multicolumn{7}{c}{\cellcolor[HTML]{EFEFEF}\textbf{Based on Mistral-7B-Instruct}} \\
    \midrule
    MEMORAG  & $37.40 \pm 1.23$ & $35.70 \pm 1.57$ & $51.20 \pm 2.45$ & $42.90 \pm 1.82$ & $30.30 \pm 1.09$ & $31.60 \pm 0.87$ \\
    TAdaRAG & $40.77 \pm 0.85$ & $39.31 \pm 1.13$ & $49.88 \pm 2.63$ & $44.83 \pm 1.55$ & $39.31 \pm 0.66$ & $36.41 \pm 0.55$ \\
    \midrule
\multicolumn{7}{c}{\cellcolor[HTML]{EFEFEF}\textbf{Based on Qwen2.5-7B-Instruct}} \\
\midrule
   MEMORAG  & $36.87 \pm 1.85$ & $36.00 \pm 1.85$ & $47.60 \pm 4.25$ & $37.99 \pm 3.45$ & $35.32 \pm 3.68$ & $31.13 \pm 2.15$ \\
    TAdaRAG & $42.38 \pm 1.33$ & $40.75 \pm 2.15$ & $46.83 \pm 3.38$ & $49.23 \pm 2.95$ & $43.79 \pm 2.95$ & $36.95 \pm 1.55$ \\
    \midrule
\multicolumn{7}{c}{\cellcolor[HTML]{EFEFEF}\textbf{Based on Qwen2.5-14B-Instruct}} \\
\midrule
   MEMORAG  & $47.37 \pm 1.05$ & $42.33 \pm 2.05$ & $55.04 \pm 3.82$ & $52.91 \pm 2.05$ & $50.42 \pm 3.31$ & $32.99 \pm 3.99$ \\
    TAdaRAG & $51.00 \pm 1.45$ & $51.32 \pm 2.09$ & $55.83 \pm 3.33$ & $56.28 \pm 2.87$ & $54.79 \pm 3.43$ & $38.36 \pm 2.73$ \\
    \bottomrule
  \end{tabularx}
  \caption{Statistical significance analysis across six datasets}
  \label{tab:stat_pub}
\end{table*}

\noindent\textbullet\hspace{0.5em} \textbf{Health dataset:} The Health dataset is derived from medical-related texts such as books, articles, and reports, covering various domains including health, medicine, diseases, drugs, and treatments. It serves as an out-of-domain dataset within the ULTRADOMAIN benchmark.

\noindent\textbullet\hspace{0.5em} \textbf{Biology dataset:} The Biology dataset is sourced from biology-related texts such as books, articles, and reports, covering various fields of biology, including cell biology, genetics, ecology, and evolutionary theory. It is an out-of-domain dataset within the ULTRADOMAIN benchmark.

\noindent\textbullet\hspace{0.5em} \textbf{Legal dataset:} The Legal dataset primarily consists of legal contracts, focusing on the legal domain and encompassing content such as contract clauses, legal terminology, and legal concepts. It is designed to test a model’s ability to understand and process the complex and nuanced language of legal documents and is one of the in-domain datasets within the ULTRADOMAIN benchmark.

\noindent\textbullet\hspace{0.5em} \textbf{HotpotQA dataset:} HotpotQA is a dataset designed to evaluate the multi-hop reasoning capabilities of question-answering systems, collected through crowdsourcing as question-answer pairs based on Wikipedia. It encompasses various types of questions and answers, such as those involving entities, dates, numbers, descriptive attributes, and "yes/no" questions comparing two entities.

\noindent\textbullet\hspace{0.5em} \textbf{2WikiMQA dataset:} 2WikiMQA is a large-scale, high-quality multi-hop question-answering dataset constructed from Wikipedia and Wikidata, designed to evaluate a model’s multi-hop reasoning capabilities. It includes four types of questions: Comparison questions, Inference questions, Compositional questions, and Bridge-comparison questions. The answer types are diverse, encompassing “yes/no,” dates, movies, people, and more.

\noindent\textbullet\hspace{0.5em} \textbf{GovReport dataset:} GovReport is a large-scale dataset comprising reports published by the U.S. Government Accountability Office and the Congressional Research Service, along with summaries written by experts, focusing primarily on the research and analysis of various policy issues. The key information in GovReport is distributed throughout the entire document, meaning that to generate accurate summaries, a model must comprehensively understand the full content of the document.

\noindent\textbullet\hspace{0.5em} \textbf{\benchmark~dataset:} \benchmark~is a real-world benchmark derived from Xinyu AI Search, featuring 3,150 multi-document QA examples in the news domain. Questions reflect real user queries on dynamic topics, with reference documents retrieved via a hybrid search engine. The dataset preserves real-world noise and redundancy, making it a challenging testbed for reasoning over fragmented, partially relevant content.

\subsection{A2: Corpus for SFT}
\label{sec:appendix A2}
\begin{figure}[t]
  \includegraphics[width=\columnwidth]{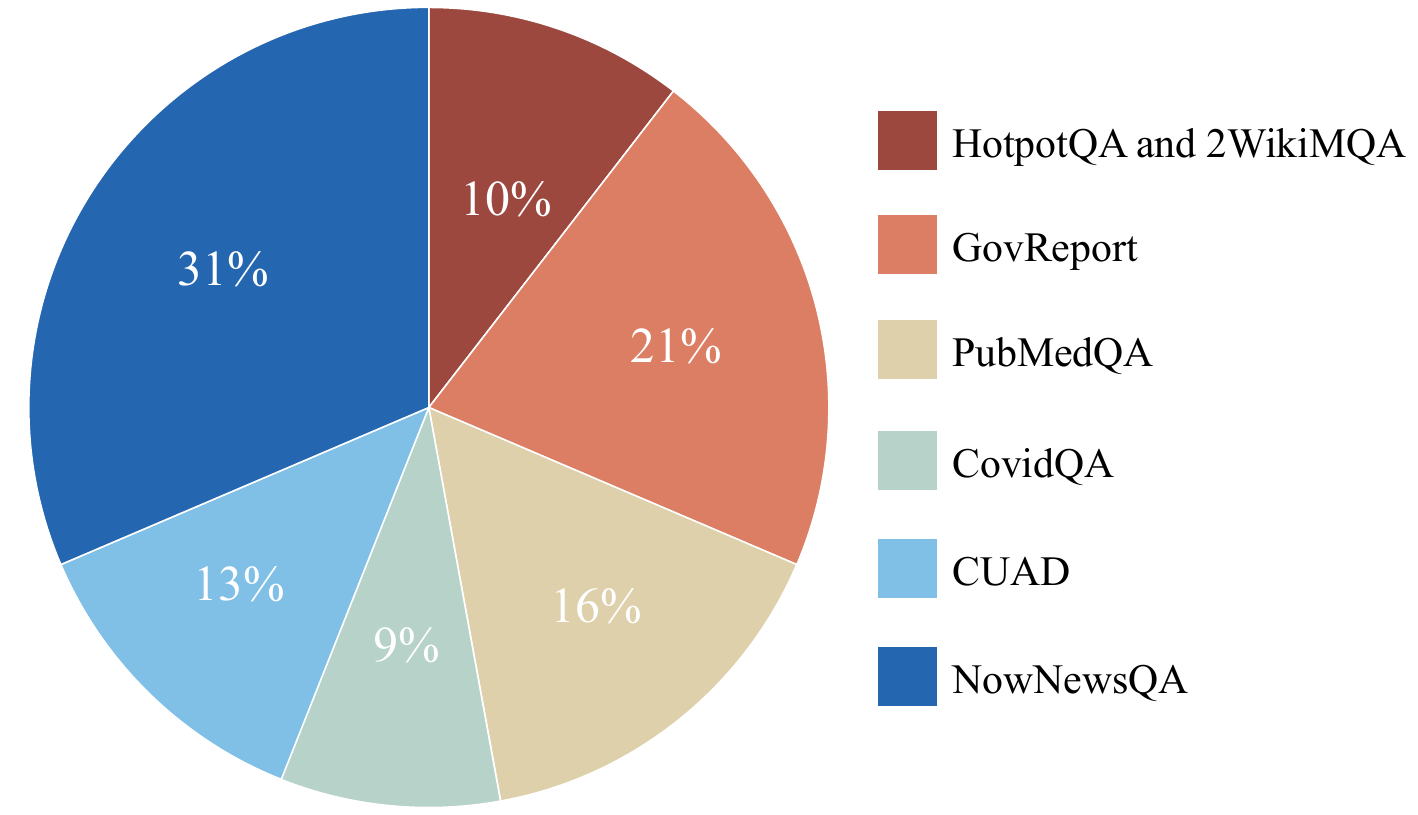}
  \caption{The distribution of SFT dataset.}
  \vskip -0.4em
  \label{fig:sft-data}
\end{figure}

We allocated the SFT dataset based on the difficulty of downstream tasks and the volume of test data, with a total data size of 9,548, as shown in Figure \ref{fig:sft-data}. For general domain tasks, some tasks (e.g., multi-document QA) require less entity information, resulting in a smaller data allocation, while others (e.g., summarization) require comprehensive information integration and high-quality entity lists, leading to a larger data allocation. In specialized domains (e.g., biomedical, legal, and news), open-ended tasks often require rich entity information from materials to improve answer accuracy and quality, leading to a higher proportion of SFT data in these domains. The introduction of each dataset can be found in Appendix A1 and Appendix A3 respectively.

\subsection{A3: Additional Datasets for SFT}
\label{sec:appendix A3}
\noindent\textbullet\hspace{0.5em} \textbf{PubMedQA dataset:}
The PubMedQA dataset originates from the PubMed database, where questions are extracted from research article titles using a rule-based heuristic approach. Each entry includes a research abstract and an automatically generated yes/no/maybe question based on the abstract's title. This dataset is highly suitable for training and evaluating text-based QA systems, particularly for question-answering applications in the biomedical domain.

\noindent\textbullet\hspace{0.5em} \textbf{CovidQA dataset:}
The CovidQA dataset is built upon the rich collection of research papers in the CORD-19 dataset, containing approximately 2,000 questions and tens of thousands of relevant background document passages. These questions and documents are not only extensive in number but also detailed in content, focusing closely on various COVID-19-related topics. This makes the dataset highly suitable for training and evaluating text-based QA systems.

\noindent\textbullet\hspace{0.5em} \textbf{CUAD dataset:}
The CUAD dataset consists of contract texts extracted from publicly available legal documents in the EDGAR database, along with questions designed by domain experts based on these contracts. It includes 21,000 questions and 510 distinct legal contract texts. This dataset is an ideal choice for training and evaluating text-based QA systems, particularly for enhancing text comprehension and question-answering capabilities in the legal contract domain.

\section{Appendix B: Baseline Descriptions}
\label{sec:appendix B}
The descriptions of the baselines are detailed as follows:

\noindent\textbullet\hspace{0.5em} \textbf{Na\"iveRAG:} This method extracts information from text databases by breaking text into smaller segments for storage, converting them into vector representations using text embeddings. When a user queries, the question becomes a vector, and the system finds the most relevant content by comparing it with stored segment embeddings based on similarity, enabling quick, direct answers as a practical RAG solution.

\noindent\textbullet\hspace{0.5em} \textbf{BGE-M3:} This method employs the novel M3-Embedding model to implement standard RAG. The M3-Embedding model uses relevance scores from various retrieval functions as teacher signals, enhancing training quality through knowledge distillation to enable joint learning and mutual reinforcement of multiple retrieval capabilities. It also optimizes batch processing strategies to support large-scale training and high throughput, thereby improving the distinctiveness of the embeddings. Furthermore, it gathers a substantial and diverse set of multilingual data from unsupervised corpora, supervised corpora, and synthetic data, providing a robust foundation for model training.

\noindent\textbullet\hspace{0.5em} \textbf{RQ-RAG:} This method can rewrite, decompose, or disambiguate user queries based on different scenarios, thereby improving the accuracy and efficiency of retrieval. It also introduces three distinct sampling strategies—based on perplexity, confidence, and ensemble approaches—to select the optimal retrieval path and answer. As a robust retrieval-augmented generation framework, it significantly enhances the knowledge retrieval and response capabilities of LLMs.

\noindent\textbullet\hspace{0.5em} \textbf{GraphRAG:} This is a graph-based RAG. It
uses an LLM to extract entities and relationships
from the text, representing them as nodes and
edges, with descriptions from the original text attached as features to reduce information loss. For
each question, a community detection algorithm
is applied to summarize and generalize the information contained in the nodes from the bottom
up, forming new community descriptions. Finally,
the results of the community detection are used to
answer global summarization questions.

\noindent\textbullet\hspace{0.5em} \textbf{HippoRAG:} This method adopts the hippocampal indexing theory from cognitive neuroscience and leverages KG techniques to enhance the retrieval capabilities of LLMs. It first converts a text corpus into a schemaless KG, where nodes represent concepts and edges denote semantic relations. For a given query, HippoRAG identifies key concepts and executes the Personalized PageRank algorithm over the KG to perform multi-hop reasoning in a single retrieval step. 

\noindent\textbullet\hspace{0.5em} \textbf{MEMORAG:} This is a novel RAG method that leverages a lightweight LLM as a memory module to compress an entire database into a compact representation known as “memory”. When a task is received, the memory module generates a series of “clues” based on this memory—rough descriptions of the answer—that guide the retrieval module in locating relevant information.

\noindent\textbullet\hspace{0.5em} \textbf{PathRAG:} This is a novel RAG approach that retrieves relevant nodes from the index graph based on keywords in the query. It then extracts key relational paths using a flow-based pruning algorithm and converts them into textual form to guide the LLM in generating more logical and coherent responses.

\begin{table*}
  \centering
  \begin{tabular}{lccccc}
    \toprule
    \textbf{Metric} & \textbf{TAdaRAG} & \textbf{PathRAG} & \textbf{$t$} & \textbf{$p$} & \textbf{$d$ [95\% CI]} \\
    \midrule
    Conciseness
      & $8.25 \pm 0.62$
      & $7.63 \pm 0.73$
      & 7.93
      & $<0.0001$
      & 0.915 [0.466, 0.774] \\
    Factuality
      & $8.45 \pm 0.53$
      & $7.85 \pm 0.55$
      & 9.62
      & $<0.0001$
      & 1.111 [0.477, 0.723] \\
    \bottomrule
  \end{tabular}
  \caption{Statistical significance analysis of \method in business scenarios with Mistral-7B-Instruct as the base model.}
  \label{tab:stat_results}
\end{table*}

\begin{figure*}[t]
  \centering
\includegraphics[width=\textwidth]{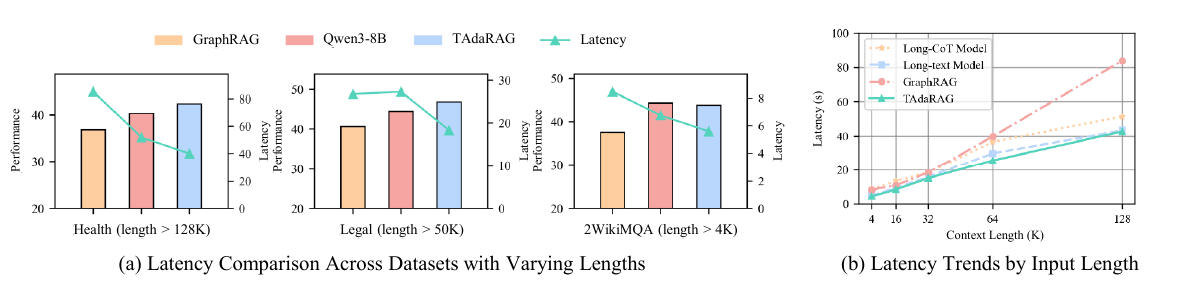} 
  \caption {Efficiency analysis results.}
  \label{fig:time}
\end{figure*}

\begin{table*}[ht]
\centering
\begin{tabular}{@{}l|cc|cc|cc|cc@{}}
\toprule
\textbf{Stage} & \multicolumn{2}{c|}{Health} & \multicolumn{2}{c|}{Biology} & \multicolumn{2}{c|}{Legal} & \multicolumn{2}{c}{HotpotQA} \\
\cmidrule(lr){2-3} \cmidrule(lr){4-5} \cmidrule(lr){6-7} \cmidrule(lr){8-9}
 & Size & Ent. & Size & Ent. & Size & Ent. & Size & Ent. \\
\midrule
Base      & 7303.12 & 58.34 & 7864.86 & 64.37 & 7287.99 & 19.97 & 1894.77 & 16.09 \\
SFT       & 5146.43 & 50.03 & 5538.78 & 55.67 & 5232.32 & 16.34 &  573.43 & 12.49 \\
Reinforce & 2005.87 & 44.23 & 2043.98 & 42.32 & 1877.84 & 15.64 &  256.67 & 10.09 \\
\midrule
\textbf{Stage} & \multicolumn{2}{c|}{2WikiMQA} & \multicolumn{2}{c|}{GovReport} & \multicolumn{2}{c|}{\benchmark} & \multicolumn{2}{c}{--} \\
\cmidrule(lr){2-3} \cmidrule(lr){4-5} \cmidrule(lr){6-7} \cmidrule(lr){8-9}
 & Size & Ent. & Size & Ent. & Size & Ent. & Size & Ent. \\
\midrule
Base      & 1524.40 & 13.32 & 6684.84 & 48.66 & 4433.23 & 31.44 & -- & -- \\
SFT       &  665.45 & 10.29 & 3077.34 & 23.73 & 2343.23 & 30.21 & -- & -- \\
Reinforce &  298.21 & 8.23  & 1240.87 & 18.82 & 1887.01 & 24.95 & -- & -- \\
\bottomrule
\end{tabular}
\caption{Average KG graph size and number of entities across datasets and training stages (Mistral-7B).}
\label{tab:kg-two-rows}
\end{table*}

\begin{figure}[t]
  \centering
\includegraphics[width=\columnwidth]{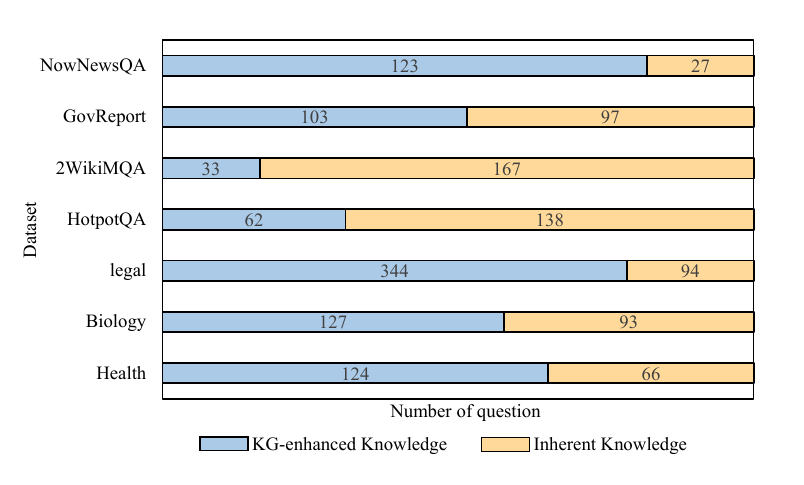} 
  \caption {Proportions of \method evidence of answers.}
  \label{fig: Evidence}
\end{figure}

\section{Appendix C: Simplified Prompt Template}
\label{sec:appendix C}

\noindent\textbullet\hspace{0.5em} \textbf{Simplified Extraction Prompt Template. }
We have carefully crafted a detailed set of extraction prompt templates. First, we guide LLMs to effectively classify questions. This framework includes six clearly defined question types specific to professional domains, each associated with a pre-established list of highly relevant entity types to ensure that the extracted entities are closely aligned with the questions. Additionally, for each specific entity type, we have customized dedicated extraction prompt templates to improve the precision and efficiency of LLMs in identifying and extracting relevant entities. For general-domain questions, we have also designed corresponding prompt templates. The simplified extraction prompt template is shown below.

\begin{figure*}[h]
\centering
\begin{tcolorbox}[colback=gray!10, 
                   colframe=black, 
                   coltitle=white, 
                   arc=2mm, auto outer arc,
                   width=\textwidth, 
                   title=Simplified Extraction Prompt Template]
\vspace{-5pt}
\textbf{List of Problem Domains and Entity Types: }\\
bio\_medical\_research: person, organism, symptom, disease, drug, technique, number, device, operation\\
general\_knowledge: person, organism, organization, location, event, time, diet, number, product\\
legal\_contracts: person, organization, location, event, time, number, contract, clause, judgment\\
customer\_support: person, technique, operation, event, time, number, device, product\\
finance: person, organization, event, time, number, operation, product, policy\\
news: person, organization, location, event, time, number, product, policy

\bigskip
\textbf{Intent Analysis: }
Given a question and a set of documents, classify the question into one of the predefined domains in {domain} by evaluating the content of the question and its relevance to the provided documents. The classification should be based on the primary focus of the question (e.g., medical, legal, financial) and the domain context found in the documents. Ensure that the answer is based on a careful assessment of the subject matter, keywords, and overall context of both the question and the document(s).\\
!! The output can only contain one domain of the problem, do not output the cause and other information !!

\bigskip
\textbf{Entities Extraction: }
Given a question, documents, and a list of entity types \{entity\_type\}, the task is to extract each type of entity step by step, and finally compile a complete list of entities. The extraction process for each entity type must strictly follow the corresponding extraction rules to identify all relevant entities in the documents that will significantly help answer the question. The output format must strictly follow the example, without any additional text or output in other formats like json. Additionally, do not output the details of entity extraction for each step, only the final list of entities.

\bigskip
\textbf{Person Entity Extraction: }
Identify all person entities relevant to answering the question. For each identified person entity, extract the following information:\\
\noindent\textbullet\hspace{0.5em} Entity Name: The name of the entity.\\
\noindent\textbullet\hspace{0.5em} Entity Type: person.\\
\noindent\textbullet\hspace{0.5em} Entity Description: A summary of the information related to the specific person in the context of the question. Based on the information, provide the most relevant and helpful description for answering the question. Optionally, include other relevant information such as the person's identity, position, major life events, significant achievements or awards, involvement in important historical events, published works, contributions, and relationships.\\
\noindent\textbullet\hspace{0.5em} Each entity's format should be \texttt{("Entity", <Entity Name>, <Entity Type>, <Entity Description>)}.

\bigskip
\textbf{Organization Entity Extraction: }
Identify all organizations relevant to answering the question. Note that the identified organizations should not include person entities! For each identified organization, extract the following information:\\
\noindent\textbullet\hspace{0.5em} Entity Name: The name of the entity.\\
\noindent\textbullet\hspace{0.5em} Entity Type: organization.\\
\noindent\textbullet\hspace{0.5em} Entity Description: A summary of the information related to the specific organization in the context of the question. Based on the information, provide the most relevant and helpful description for answering the question. For example, the extracted entity description may include the organization's name, function, goals, leadership, historical background, culture, and its role or influence in the related events. For corporate organizations, in addition to extracting the company name and function, also focus on financial data, market share, annual revenue, number of employees, and other numerical information.\\
\noindent\textbullet\hspace{0.5em} Each entity's format should be \texttt{("Entity", <Entity Name>, <Entity Type>, <Entity Description>)}.

\bigskip
......
\vspace{-5pt}
\end{tcolorbox}
\end{figure*}

\noindent\textbullet\hspace{0.5em} \textbf{Multi-Faceted Evaluation Criteria. } 
The detailed definitions of the multi-faceted evaluation criteria are provided below. The original text is in Chinese, and we translate it into English to align with the language of this paper.

\begin{figure*}[h]
\centering
\begin{tcolorbox}[colback=gray!10, 
                   colframe=black, 
                   coltitle=white, 
                   arc=2mm, auto outer arc,
                   width=\textwidth, 
                   title=Multi-Faceted Evaluation Criteria]
\vspace{-5pt}
\textbf{"Relevance":} "The response should directly address the user's question, avoiding irrelevant content, unnecessary elaboration, or roundabout explanations. Each irrelevant statement deducts one point.",

\textbf{"Numerical Precision":} "For questions asking for specific numbers, avoid vague responses like 'several' or 'many.' Answers should be precise and specific. Each ambiguous response deducts one point.",

\textbf{"Conciseness":} "If the user specifies a time period, location, person, or event, the answer must meet these requirements. Each statement that fails to address the question properly deducts one point.",

\textbf{"Factuality":} "Information in the response must be accurate, especially for factual questions. Avoid incorrect numerical data or factual errors. Each numerical or factual mistake deducts one point.",

\textbf{"Timeliness":} "For ongoing news events or urgent reports, the provided information should reflect the latest updates. Note that today's date is xxxx-xx-xx. If the question is not time-sensitive, no points will be deducted. For time-sensitive questions, evaluate the response’s timeliness; the less timely it is, the more points should be deducted accordingly.",

\textbf{"Comprehensiveness":} "The response should comprehensively cover all aspects of the user's inquiry, providing detailed information. The answer should be self-contained, requiring no further search to understand the full context. Each missing essential point deducts one point.",

\textbf{"Clarity":} "The response should be clear and easy to understand, with a well-structured format for readability. It should quickly convey the necessary information. (Example of unclear expression: Using a parallel or topic-based structure instead of a chronological one when a time-sequence approach is clearly needed.) Each unclear expression deducts one point.",

\textbf{"Coherence":} "The response should be logically coherent, with smooth transitions between sentences and appropriate word choices. Each instance of incoherence deducts one point.",

\textbf{"Insightfulness":} "The response should provide unique insights and depth. The base score for this criterion is 6 points, with each innovative viewpoint or expression earning an additional 0.5–1 point."

\bigskip

\vspace{-5pt}
\end{tcolorbox}
\end{figure*}

\begin{figure*}[h]
\centering
\begin{tcolorbox}[colback=gray!10, 
                   colframe=black, 
                   coltitle=white, 
                   arc=2mm, auto outer arc,
                   width=\textwidth, 
                   title=Prompt for multi-faceted evaluation]
\vspace{-5pt}

Assume you are an article quality inspector. Please evaluate the response based on \{ Metric Title \}. I will provide the user's question and the final response. The maximum score is 10 points, and the scoring rules are as follows:

\medskip 
\{ Metric Definition \}

\medskip

Please strictly follow the scoring rules. Example output format:

\medskip

\texttt{\textcolor{black}{\{}
\begin{flushleft}
\small
" Issues Identified ": " X ", \\
" Calculation Process ": "10 -1.0 -1.0 -1.0 = 7.0", \\
" Score ": 7
\end{flushleft}
\textcolor{black}{\}}}

\medskip

\{ Few - Shot Examples \}

\medskip

Your final score: \hfill "

\bigskip

\vspace{-5pt}
\end{tcolorbox}
\end{figure*}

\section{‌Appendix D: Additional Results}
\label{sec:appendix E}
\subsection{D1: Statistical Significance Analysis }
\label{sec:appendix E1}
\textbf{For Public Dataset.} We performed statistical significance tests on \method and MEMORAG across six public benchmarks (Health: \(n=220\); Biology: \(n=180\); Legal: \(n=438\); HotpotQA: \(n=200\); 2WikiMQA: \(n=200\); GovReport: \(n=200\)). As shown in Table~\ref{tab:stat_pub}, take mistral-7B as backbone, for Health, Biology, 2WikiMQA, and GovReport, the lower bound of \method’s 95\% CI exceeds the upper bound of MEMORAG’s 95\% CI, indicating significant improvements (\(p<0.001\)). For Legal and HotpotQA, MEMORAG’s means are slightly higher but their CIs overlap, indicating non‑significant differences (\(p>0.05\)).

\textbf{For Business Scenarios.}  
We conducted independent t-tests on \method and PathRAG using 300 samples per metric to assess performance differences, as shown in Table \ref{tab:stat_results}.
For \textit{conciseness}: TAdaRAG (M=8.25, SD=0.62) significantly outperformed PathRAG (M=7.63, SD=0.73), $t(298) = 7.93$, $p < 0.0001$, $d = 0.916$, 95\% CI = [0.667, 1.165].
For \textit{factuality}: TAdaRAG (M=8.45, SD=0.53) also showed a significant advantage over PathRAG (M=7.85, SD=0.55), $t(298) = 9.62$, $p < 0.0001$, $d = 1.111$, 95\% CI = [0.877, 1.344].
These results highlight the superior performance of \method in business contexts, particularly in generating concise and factually accurate outputs.

\subsection{D2: Efficiency Analysis}
\label{sec:appendix E2}
We conducted an efficiency analysis on three datasets with significant length differences, as shown in Figure~\ref{fig:time}(a). The results indicate that compared to the Long-CoT model and the classic GraphRAG method, our model, considering both dynamic graph construction time and first-token generation time, demonstrates superior latency performance. Additionally, it exhibits stable and outstanding performance across datasets of varying scales. In particular, compared to GraphRAG, which relies on pre-built index graphs and subgraph retrieval, our approach achieves significant improvements in both efficiency and effectiveness.

Furthermore, we analyzed the latency trends of different models as the input document length varies in practical tasks, shown in Figure~\ref{fig:time}(b). The experimental results demonstrate that compared with various long-context answering methods, our model consistently maintains the lowest latency across all input lengths, further validating its strong adaptability to diverse input scales and its acceptable response time cost in real-world applications.

\subsection{D3: KG Optimization Results}
To evaluate the effectiveness of reinforcement learning-based knowledge graph optimization (RL-based KG optimization), we analyzed the knowledge graph size (Size) and entity number (Ent.) before and after training across seven datasets, as shown in Table~\ref{tab:kg-two-rows}. The results indicate a significant reduction in graph size and entity count after training. For example, the Health dataset's graph size decreased from approximately 7,300 to around 2,000, and the entity number dropped from 58 to 44, demonstrating substantial compression.

These changes suggest that the model successfully learned to control the graph size by effectively removing redundant and unnecessary information, focusing on core entities and relations, thereby improving the quality and efficiency of knowledge graph representation and reasoning. The optimization effect is especially notable in large-scale datasets such as Health and GovReport, providing more accurate and efficient knowledge support for downstream tasks.

\subsection{D4: Evidence Analysis}
We analyzed the answers across seven datasets to investigate the sources of evidence supporting \method in generating answers, as illustrated in Figure~\ref{fig: Evidence}.
The results indicate that in the general domain, which includes a considerable amount of commonsense questions, a significant proportion of the answers rely exclusively on the intrinsic knowledge embedded within LLM’s parameters.
However, in specialized domains such as the medical domain and legal domain, accurate answer generation requires specific domain knowledge.
Consequently, the experimental results demonstrate that a substantial proportion of these queries rely on a combination of knowledge from both retrieved KG and the LLM’s inherent knowledge for answer generation.


\section{Appendix E: Case Study}
\begin{figure*}[t]
  \includegraphics[width=\textwidth]{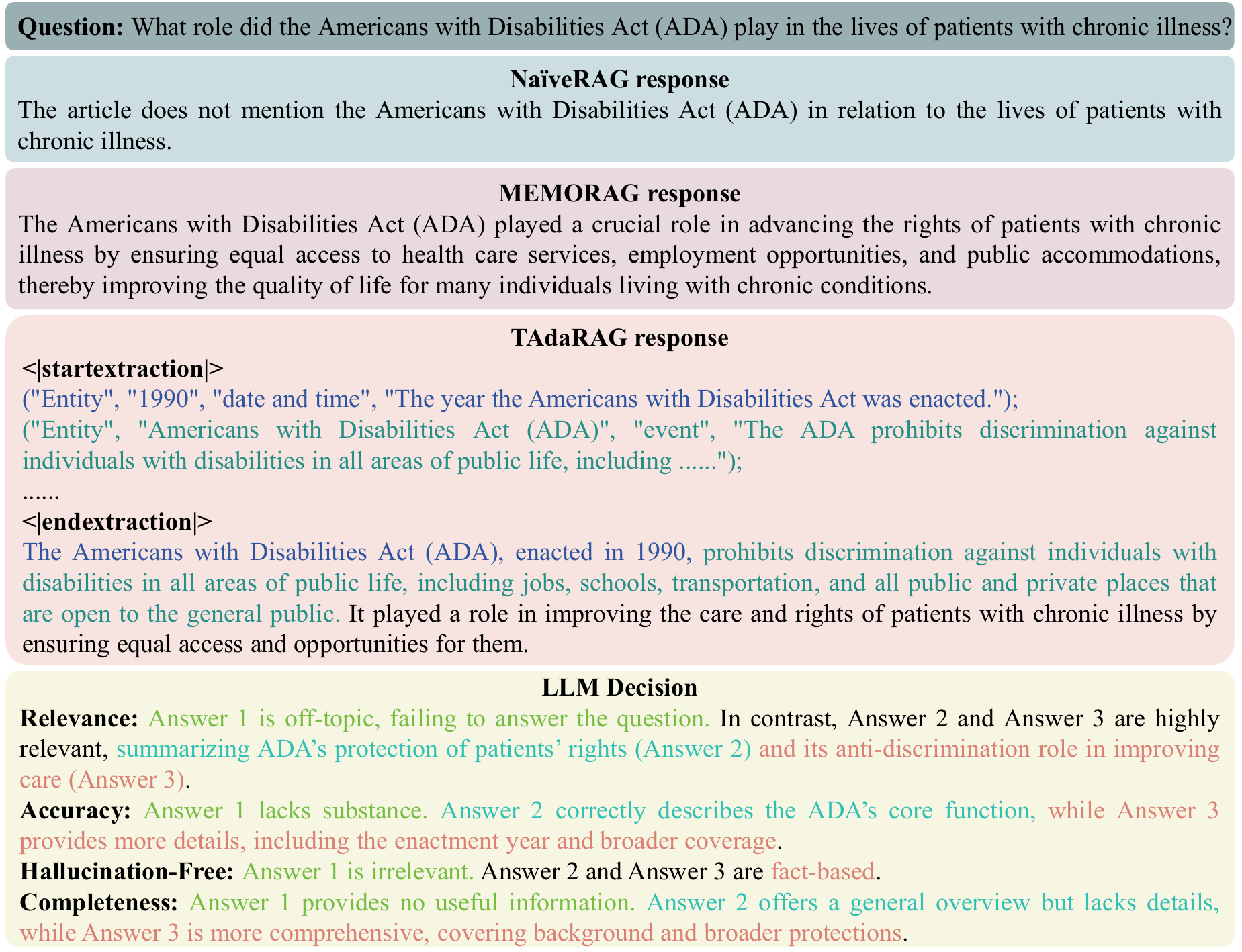}
  \caption{The figure displays the answers provided by different methods to the same question in the Health dataset, along with the LLM's analysis of these answers, visually demonstrating the differences in answer quality across the various methods.}
  \label{fig:case}
\end{figure*}

\begin{figure*}[t]
  \includegraphics[width=\textwidth]{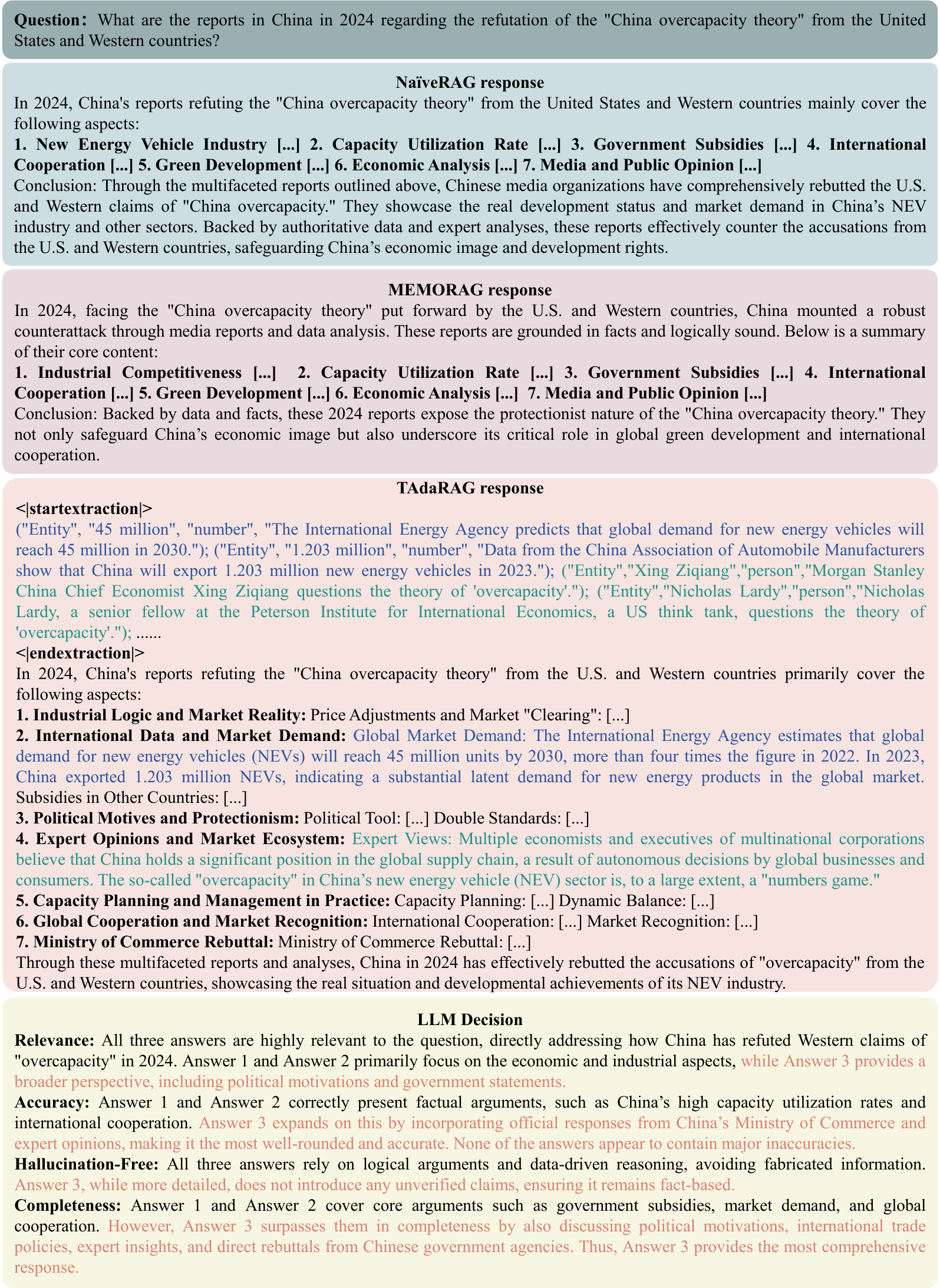}
  \caption{The figure displays the answers provided by different methods to the same question in the \benchmark~dataset, along with the LLM's analysis of these answers, visually demonstrating the differences in answer quality across the various methods.}
  \label{fig:new_case}
\end{figure*}

We additionally provide two case studies, as shown in Figure \ref{fig:case} and Figure \ref{fig:new_case}, demonstrating the different responses generated by Na\"iveRAG, MEMORAG, and \method to the same question using the Health and \benchmark~datasets with Mistral-7B-Instruct as the base model. These responses were evaluated by GPT-4o across four dimensions: relevance, accuracy, hallucination-free performance, and completeness. The results indicate that the charts extracted through our method significantly enhance the quality of the responses. Compared to other approaches, our method exhibits a clear advantage in response performance.


\end{document}